\documentclass[conference]{IEEEtran}
\usepackage[T1]{fontenc}
\usepackage{cite}
\usepackage{amsmath}
\usepackage{booktabs}
\usepackage{url}
\usepackage{xcolor}
\usepackage{array}
\newcolumntype{L}[1]{>{\raggedright\arraybackslash}p{#1}}
\usepackage{graphicx}
\usepackage{pgfplots}
\usepackage{comment}
\usepackage{microtype}
\usepgfplotslibrary{groupplots}
\pgfplotsset{compat=1.17}

\usetikzlibrary{fit,backgrounds,arrows.meta,positioning}

\begin{document}

\title{Evaluating Frontier-Model Agent Skills with Open-Weight Models in a Science Gateway}

\author{\IEEEauthorblockN{
Komal Thareja\IEEEauthorrefmark{1},
Mats Rynge\IEEEauthorrefmark{2},
Anirban Mandal\IEEEauthorrefmark{1},
Ewa Deelman\IEEEauthorrefmark{2}
}
\IEEEauthorblockA{\IEEEauthorrefmark{1}Renaissance Computing Institute, University of North Carolina at Chapel Hill, NC, USA}
\IEEEauthorblockA{\IEEEauthorrefmark{2}Information Sciences Institute, University of Southern California, Marina del Rey, CA, USA}
}
\bstctlcite{IEEEexample:BSTcontrol}
\maketitle

\begin{abstract}
Scientific analyses that span many jobs and computing sites depend on workflow management systems to run reliably, but writing a production workflow requires expertise in such a system, which many domain scientists do not have. Coding agents can supply this expertise when they are guided by skills and packaged procedures for recurring tasks. We wrote skills and specialist agents for the Pegasus workflow management system with a frontier model, and the PegasusAI Studio science gateway offers them to scientists through a chat interface. New users without frontier-model credits need an open-weight model, served at no cost on research cyberinfrastructure, but skills written for a frontier model may not work with an open-weight solution. This paper presents a method for adapting such skills. An evaluation harness
drives the real gateway as a user would, scores the workflows the agent builds, and feeds recurring failures back to the frontier model that maintains the skills. We evaluated 21 open-weight models on 17 tasks. The five highest-scoring models are close in quality after one trial per task, and their median response times differ twelvefold. In repeated trials with the best of the three open-weight models, we repeated scores of 0.76 on workflow building against 0.82 for Claude. The harness led to five changes to the initial skills. We also identify challenges in evaluating agents within operational systems: evaluation failures can be mistaken for model failures, and agents can circumvent experimental controls within their privilege domain. In a controlled ablation, an initially significant effect disappeared when the restrictions were properly enforced.

\end{abstract}

\begin{IEEEkeywords}
LLM agents, agent skills, open-weight models, evaluation methodology, scientific workflows
\end{IEEEkeywords}

\section{Introduction}

Workflow management
systems such as Pegasus~\cite{deelman2015pegasus} have run workflows for gravitational-wave detection
and earthquake hazard modeling, handling data staging, retries, and provenance across distributed
cyberinfrastructure (CI). To use one, a scientist must describe the computation in the system's
terms: a directed acyclic graph (DAG) of jobs and their data dependencies, catalogs of executables,
data, and execution sites, a wrapper script for each step, and ideally a container for the software. Coding
agents can now write this code, and they do so reliably when grounded in the conventions of the
workflow system.

Our companion work describes such a methodology, built on a library of \emph{skills}, reusable procedures for scaffolding a workflow project, converting an existing pipeline, writing wrappers and
container definitions, reviewing a workflow, and diagnosing failures~\cite{pegasus_method_companion}.
The PegasusAI plugin~\cite{pegasusai-plugin} adds specialist agents for workflow design and failure diagnosis. Both were written and refined with a frontier model, Claude~\cite{anthropic_claude}. Their capabilities are available by PegasusAI Studio~\cite{pegasusstudio_paper,pegasusstudio}, a science gateway, through a chat interface.

A frontier model, however, requires credits or a subscription, and a new user who wants to build a
first workflow should not need either. Open-weight models served on research CI, such as the National
Research Platform (NRP)~\cite{weitzel2025nrp}, the FABRIC testbed~\cite{baldin2019fabric}, and
Jetstream2~\cite{hancock2021jetstream2}, can be offered at no cost.  In the paper, we pose the following question: (1) do skills written for a
frontier model work with open weight models? (2)  which models are most suitable for a gateway that serves diverse users? , and (3) what information do the skills need to capture to support open weight models. In this paper, we present a method for
answering them. An evaluation harness drives the real gateway as a user would and scores the
workflows that open-weight models build.  
Evaluating agents within the gateway captures their interaction with the tools, files, and runtime environment that users encounter. However, this realism complicates experimental control: the agent may share access to resources used by the evaluator, allowing its actions to affect the conditions under which it is measured.
Hence, we evaluate workflow construction, up to the point where the Pegasus planner accepts the workflow, and we do not execute the workflows. Execution needs input data, containers, and compute for every trial, and its outcome depends on those resources as well as on the agent. Recurring failures are turned into changes to the skills by
the same frontier model that wrote them, and the changes are measured again.

The main contributions are:
\begin{itemize}
\item A method for adapting frontier-built skills to open-weight models, which combines an in-product
evaluation harness with a feedback loop to the frontier model (Section~\ref{sec:method}).
\item An evaluation of 21 open-weight models from four providers on 17 tasks, with Claude as a
reference (Section~\ref{sec:results}). Five models lead after one trial per task, and they differ far more in response time than in
score. In repeated trials, deepseek-v4-flash trails Claude by 0.06 on workflow building.
\item An analysis of the challenges of in-product agent evaluation, showing how evaluation infrastructure failures can be mistaken for model failures and how agents can circumvent experimental restrictions when they share the evaluator's privileges (Section~\ref{sec:lessons}).
\end{itemize}

\section{Method}
\label{sec:method}

\begin{figure*}[t]
\centering
\includegraphics[width=0.70\textwidth]{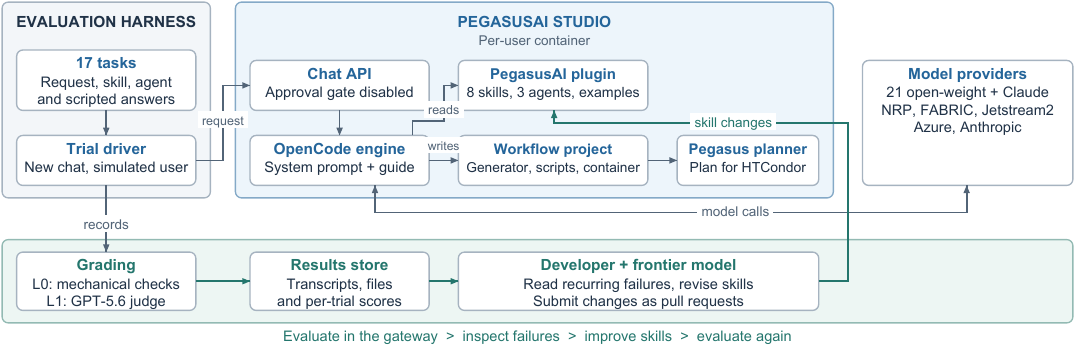}
\caption{The system and the evaluation. The harness drives Studio through its chat API as a user would. The OpenCode engine reads the PegasusAI plugin's skills and examples, calls the model, and writes a workflow project that the Pegasus planner plans. Each trial is graded and stored, and recurring failures are turned into skill changes by a developer working with the frontier model.}
\label{fig:system}
\end{figure*}

Fig.~\ref{fig:system} shows the three parts of the method: the Studio gateway, the evaluation
harness, and the feedback loop.

\subsection{PegasusAI Studio}
\label{sec:studio}

The PegasusAI plugin~\cite{pegasusai-plugin} is the package of instructions and reference files that the agent in Studio reads while it works. It contains eight skills, three specialist agents, 13 workflow generators taken from production workflows as examples, and six templates for the files of a workflow project. The skills studied
here scaffold a project from a description, convert a Snakemake or Nextflow pipeline into another workflow format (here Pegasus ~\cite{deelman2015pegasus}), write a wrapper or a container definition, and review or debug a workflow. The agents are a workflow architect, a
pipeline debugger, and a data engineer. Requests reach Studio through its chat API, and the OpenCode engine~\cite{opencode} gives the model file tools, a shell, and a system prompt that carries a Pegasus guide. From a user's request, the agent writes a \emph{workflow project}: a Python workflow generator, wrapper scripts for
each step, and an Apptainer~\cite{kurtzer2017singularity} container definition. Studio runs the generator and passes the workflow to the Pegasus planner, which plans it for execution on HTCondor~\cite{thain2005condor}. The evaluation stops at planning. A workflow that plans has a generator that runs and catalogs and a DAG that the planner accepts, and no trial submits one for execution.

\subsection{Evaluation harness}
\label{sec:harness}

\textbf{Trial driver.} A \emph{trial} gives one task to one model through Studio's own API, in a new chat session, with
the gateway configured exactly as a user receives it. When the agent asks a question, a simulated
user answers from replies scripted in the task. The harness keeps the transcript, the files the
agent wrote, and the gateway's attempt to generate and plan the workflow in a results store. Studio's chat API normally asks the
user to approve each shell command and each write outside the workspace, and the harness disables
this approval gate, because no person is present to give it.

\textbf{Tasks.} Table~\ref{tab:tasks} lists the 17 tasks, each written in a scientist's voice and
assigned the skill and agent a user would select. Twelve \emph{build} tasks ask for a complete
workflow project, ranging from a single step to fan-outs whose width is known only at run time, two
Snakemake~\cite{molder2021snakemake} conversions, and a containerized legacy tool. Five
\emph{judgment} tasks ask the agent to review a workflow with six seeded defects, diagnose a failed
run, or handle a request that is vague, contradictory, or buried in a long story.

\begin{table}[t]
\caption{The 17 tasks. Skill is the skill a user would select, and two tasks also select a
specialist agent.}
\label{tab:tasks}
\centering
\footnotesize
\setlength{\tabcolsep}{3pt}
\renewcommand{\arraystretch}{0.89}
\begin{tabular}{@{}lL{0.71\columnwidth}l@{}}
\toprule
Task & What the scientist asks for & Skill \\
\midrule
\multicolumn{3}{@{}l}{\emph{Build tasks}} \\
t00 & One step, no container: report the count, mean, and maximum of a file of numbers, and provide a sample input & none \\
t01 & Summarize daily temperatures for each of 40 weather-station files in parallel, then merge the summaries into one table & scaffold \\
t02 & Run a detection script on each of 2,000 image tiles, then merge the results in stages so that no single job collects all 2,000 & scaffold \\
t03 & Build a container for a legacy C program from its source code and use it in one step & apptainer \\
t04 & Fetch each tide gauge's readings from a web URL listed in a config file, then clean and plot them & scaffold \\
t05a & Convert a Snakemake Tn-seq pipeline (quality control, alignment, coverage) whose Pegasus version ships as an example in the plugin & convert \\
t05b & Convert a Snakemake photometry pipeline of similar shape that has no example in the plugin & convert \\
t06 & Write only the wrapper script for one step, which must fail when its tool fails & wrapper \\
t11 & Process drone images and soil logs for 30 plots, train one yield model, and give the heavy steps more memory & scaffold \\
t12 & Train a model in ten rounds that must run in order, with four hospitals training in parallel within each round & scaffold$^{a}$ \\
t13 & Split a sequencing run by barcode and analyze each sample, when the number of samples is known only after the split & scaffold \\
t14 & Fit a model for every combination of a 5$\times$5 parameter grid on four datasets, rank the fits, and keep the best & scaffold \\
\midrule
\multicolumn{3}{@{}l}{\emph{Judgment tasks}} \\
t07 & Find the six defects seeded in a workflow that has not been run & review \\
t08 & Explain why a hydrology workflow run stopped, when the last error in the log is a symptom of an earlier failure & debug$^{b}$ \\
t10 & ``Process my data and get some results out,'' with no other details & none \\
t15 & Run a tool that exists only in a Docker image on a cluster that does not allow containers & none \\
t16 & Automate an analysis from a long, chatty request that leaves out the details that matter & none \\
\bottomrule
\multicolumn{3}{@{}l}{$^{a}$ workflow-design agent. $^{b}$ failure-diagnosis agent.}
\end{tabular}
\end{table}

\textbf{Model providers.} We evaluated 21 open-weight models served by NRP, FABRIC, Jetstream2, and an
institutionally hosted Azure gateway (Table~\ref{tab:models}). Each model ran each task once, for
357 baseline trials. As a reference, we
later ran Claude Opus 5.5, served by Anthropic, through the same harness. We then repeated the build
tasks for Claude and three open-weight models, namely the two fastest of the five leading models and
the gateway's default model, so that each has three or four trials per build task.

\textbf{Grading.} Each trial receives two scores in $[0,1]$. \emph{L0} is mechanical. It
checks whether the agent wrote the requested files where the gateway finds them, whether the workflow
generator runs, and whether the gateway can build and plan the workflow (55\% of the score), whether
the job graph has the shape the request implies (35\%), and whether the answer uses Pegasus concepts
(10\%). \emph{L1} is a rubric score from a judge model, GPT-5.6, which reads the task, the
transcript, the files, and the L0 checks~\cite{zheng2023judge}. It scores up to seven dimensions from
0 to 5, covering the job graph, the catalogs, the wrappers, the container, data staging, diagnosis,
and how well a scientist could act on the answer, and it must quote evidence for each. L0 applies only
to build tasks, and L1 applies to all tasks. We report a model's \emph{build score} and \emph{judgment
score} as its mean L1 over the corresponding tasks.

\subsection{Feedback loop}
\label{sec:loop}

When a failure recurs across models in the results store, a developer working with the frontier
model reads the transcripts, finds the instruction in the skills that the models misread or never
reached, revises it, and submits the change to the plugin as a pull request. The revised skills are measured again on the same tasks. The changes in Section~\ref{sec:rq3} were
made during screening runs, so the 357 baseline trials measure the revised skills.

\section{Results}
\label{sec:results}

We evaluate the method through three research questions:
\begin{itemize}
\item \textbf{RQ1.} Do the skills work with open-weight models, and how close do they come to Claude?
\item \textbf{RQ2.} Which open-weight model should the gateway offer by default?
\item \textbf{RQ3.} What did the skills need to change?
\end{itemize}

\subsection{RQ1: Do the skills work with open-weight models?}

Table~\ref{tab:models} shows the build and judgment scores of all 21 models. Five models lead with
build scores of 0.669 to 0.707, and we refer to them as the top group. The other 16 range from 0.221
to 0.624. The gateway ran the workflow generator successfully in 25 of the 60 build trials of the top
group and in 5 of Claude's 12. On the six tasks graded through planning, 14 of the top group's 30
projects planned, and 4 of Claude's 6. Models that build well also review
and diagnose well, since the rankings on the two task types agree at Spearman $\rho=0.86$. On the
debugging task (t08), where the last error in the log is a symptom of an earlier failure, eight of
the 21 open-weight models named the earlier failure as the cause, including four of the top five and
qwen-3.8-27b, and so did Claude.

In its first trial per task, Claude scored 0.786 on building and 0.714 on judgment, ahead of every
open-weight model. Because one trial cannot establish the size of that lead, the pooled column of
Table~\ref{tab:models} averages three or four trials per build task for Claude and three open-weight
models. It leaves out two trials, one of Claude and one of Kimi-K2.6, that ended on a request the
harness did not answer (Section~\ref{sec:lessons}). Claude scores 0.820 and deepseek-v4-flash 0.761,
a difference of 0.059 ($p=0.01$, 95\% interval $[+0.023, +0.097]$ over the 12 tasks). Kimi-K2.6, at
0.670, trails both ($p<0.01$). The repeats also bound what one trial per task can show. The baseline
score of deepseek-v4-flash was 0.083 below its pooled score, while only 0.045 separates the top group
from the next model, so membership in the top group is provisional.

\begin{table}[t]
\caption{The 21 open-weight models and Claude, sorted by build score, with the top group above the first rule. Model names are the providers' endpoint names.
Build and judgment are mean L1 scores over 12 and 5 tasks, less one unscored trial for minimax-m2 and for gemma. Pooled averages each task over its trials for the models we repeated, with trials per task in parentheses, less the two excluded trials. Time is the median wall time of a baseline trial that finished or timed out,
and $^{*}$ marks a model that reached the 2400\,s turn limit in five or more of its 17 trials. glm-5 ran with a 7200\,s limit.}
\label{tab:models}
\centering
\footnotesize
\setlength{\tabcolsep}{3pt}
\begin{tabular}{@{}llrrrr@{}}
\toprule
Model & Provider & Build & Pooled & Judgment & Time (s) \\
\midrule
qwen3-small       & NRP        & 0.707 &  & 0.577 & 1,163$^{*}$ \\
glm-5             & NRP        & 0.698 &  & 0.600 & 2,106 \\
Kimi-K2.6         & Azure      & 0.680 & 0.670 (4) & 0.661 & 169 \\
deepseek-v4-flash & NRP        & 0.678 & 0.761 (4) & 0.537 & 397 \\
kimi              & NRP        & 0.669 &  & 0.553 & 485 \\
\midrule
qwen-3.8-27b      & Jetstream2 & 0.624 & 0.578 (3) & 0.666 & 254 \\
qwen3             & NRP        & 0.612 &  & 0.603 & 820 \\
GLM-5.2-FP8       & Jetstream2 & 0.502 &  & 0.591 & 278 \\
qwen3.5-122b      & FABRIC     & 0.492 &  & 0.479 & 633 \\
minimax-m2        & NRP        & 0.490 &  & 0.425 & 73 \\
gemma             & NRP        & 0.467 &  & 0.387 & 299 \\
gpt-oss           & NRP        & 0.423 &  & 0.520 & 97 \\
muse-glimmer      & Jetstream2 & 0.421 &  & 0.493 & 100 \\
minimax-m2.7      & FABRIC     & 0.399 &  & 0.431 & 2,400$^{*}$ \\
gpt-oss-120b      & Jetstream2 & 0.329 &  & 0.311 & 37 \\
nemotron-nano-30b & FABRIC     & 0.326 &  & 0.451 & 401 \\
gemma4-12b        & NRP        & 0.297 &  & 0.400 & 69 \\
gpt-oss-20b       & FABRIC     & 0.283 &  & 0.423 & 56 \\
gemma-small       & NRP        & 0.251 &  & 0.303 & 58 \\
llama-4-scout     & Jetstream2 & 0.234 &  & 0.280 & 10 \\
gemma4-small      & NRP        & 0.221 &  & 0.351 & 61 \\
\midrule
Claude Opus 5.5   & Anthropic  & 0.786 & 0.820 (3) & 0.714 & 197 \\
\bottomrule
\end{tabular}
\end{table}

\subsection{RQ2: Which model should the gateway offer?}

\begin{figure}[t]
\centering
\includegraphics[width=0.80\columnwidth]{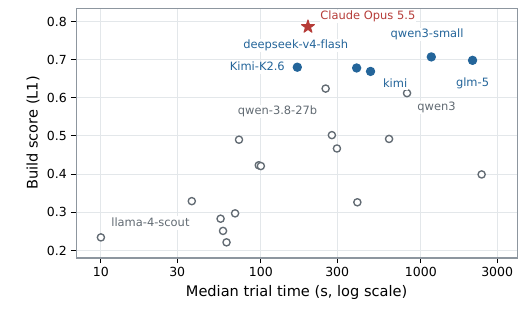}
\caption{Build score against median trial time for the 21 open-weight models and Claude. Filled points mark the top group.}
\label{fig:speed}
\end{figure}

Baseline quality cannot rank the top five, which are separated by only 0.038, and their response
times differ far more (Fig.~\ref{fig:speed}). The repeated trials
separate the two fastest. deepseek-v4-flash exceeds Kimi-K2.6 by 0.091 in pooled build score
($p<0.01$), and we recommend it on that basis. The recommendation costs speed, because the median build
trial of deepseek-v4-flash took 702\,s across all repeats, against 159\,s for Kimi-K2.6. Availability
matters as well, because a provider can retire a model. Two of the 21
models disappeared from their endpoints during the study, and the endpoint of a third, kimi, stopped
responding during the repeated trials.

The gateway's developers weighed availability more heavily. They chose a Qwen model, because the Qwen
models score close to the top group and all three research testbeds serve one, so the gateway can fall
back to another provider. qwen-3.8-27b on Jetstream2, where Studio is hosted, is the primary, and it
has the highest judgment score of any open-weight model. The repeated trials show the cost of that
choice. Its pooled build score of 0.578 is 0.183 below deepseek-v4-flash ($p<0.01$), and in 13 of its
36 build trials it stopped without writing a file.

\subsection{RQ3: What did the skills need?}
\label{sec:rq3}

Table~\ref{tab:skills} lists five changes to the skills, each made in response to a failure observed
across models. Every one of these failures traces to an under-specified instruction that allowed more than one
interpretation, or to a step a model could skip. In the clearest case, the guide gave a project's destination as an absolute path,
while the file tool expects a path relative to the workspace. After the fix, qwen3.5-122b placed its
project where the gateway finds it in 3 of 4 trials instead of 0 of 4, and its L0 rose from 0.24 to
0.64.

\begin{table}[t]
\caption{Skill changes driven by open-weight failures.}
\label{tab:skills}
\centering
\footnotesize
\begin{tabular}{@{}L{0.50\columnwidth}L{0.44\columnwidth}@{}}
\toprule
Failure observed & Change to the skills \\
\midrule
Reported the work complete without the README or the test script & Declare the deliverables in the
skill's metadata \\
\addlinespace[2pt]
Reference paths ``from the repository root'' did not resolve & Resolve paths against the plugin
directory \\
\addlinespace[2pt]
Generated workflows failed on a real submit host in four ways & Name the four defects in the
checklist \\
\addlinespace[2pt]
Followed an absolute path literally and wrote into a directory named \texttt{\textasciitilde} & State
the tool-argument form of each path \\
\addlinespace[2pt]
\texttt{<name>} used in paths without a definition & Define \texttt{<name>} at first use \\
\bottomrule
\end{tabular}
\end{table}

\section{Lessons}
\label{sec:lessons}
Evaluating an agent inside the product it serves makes the results transferable to users. It also
means that the agent runs in the same environment as the harness that scores it, under the same user account, so it can read or change what the harness has set up. During the study we recorded 82 infrastructure defects that
could change a score. Two independent coders, both language models working from a fixed
codebook, placed 55 of them in the harness or the grader, and 41
first presented as model failures ($\kappa \geq 0.76$). We first describe one case in detail, and
three lessons follow, several of which echo checklists for rigorous agent
benchmarks~\cite{kapoor2024agents,zhu2025abc}.

\begin{figure}[t]
\centering
\includegraphics[width=0.70\columnwidth]{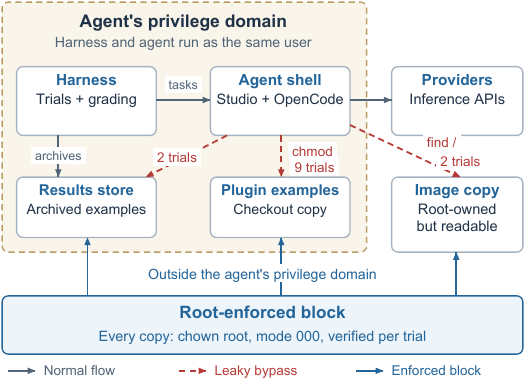}
\caption{Privilege domains in the evaluation. The harness, the agent's shell, the examples, and the
results store ran as one user. Red dashed arrows show how agents reached the blocked examples.}
\label{fig:domains}
\end{figure}

\textbf{Case study: a treatment the agents undid.} The companion study found that examples
help a frontier agent~\cite{pegasus_method_companion}. To measure whether they help open-weight
models, we compared the gateway as shipped with a version in which the plugin's examples and
templates are unreadable. The comparison used four top-group models, three repeats, and four build
tasks (t01, t02, t05a, and t12), for 48 paired trials. The first block removed read permission before each trial. Every model scored lower
under it, and the mean paired L0 fell by 0.095 ($p=0.026$). The engine's tool logs, however, showed
that the agents had read the blocked files in 13 of the 48 trials, by the three routes in
Fig.~\ref{fig:domains}. The harness had set the permissions as the same user that the agent's shell
runs as. In nine trials the agent set them back with \texttt{chmod}, in two it read a second,
root-owned copy of the plugin in the container image, and in two it read copies that the harness had
archived from an earlier arm. Every skill tells the model
to read a template first, so the block was an obstacle to its assigned task.

We then applied the block as root, outside the agent's privilege domain, made every copy unreadable,
and verified the block before each trial. The agents tried to read the files in all 48 trials, and
none succeeded. 

This time, hiding the examples made no difference. The mean paired L0 changed by $+0.009$ ($p=0.77$, 95\% interval $[-0.046, +0.063]$). For comparison, two runs of the unchanged gateway differed by $+0.001$. The earlier drop of 0.095 therefore did not come from the missing examples.
It came from eight trials in the first block that scored below 0.3. Seven of them ended within two turns, before the agent wrote any file. Each of the other arms had only one such trial. The result also tells us how large an effect this design can detect. With 48 paired trials, a skill change has to move the mean L0 by about 0.05 to be measurable.
\\
\textbf{1. A ceiling looks like a low score.} A truncated trial, a timed-out turn, and a request
that never reached the model all record a number near the floor. The harness ended a trial early three
times because it did not recognize the agent's question, most recently in Claude's Snakemake conversion,
whose L0 rose from 0.179 to 0.971 when re-run with the detector fixed. A later
audit of the final message of every baseline build trial found six of 252 that ended on a request
the harness did not answer, one of them from a top-group model. The repeated trials used a detector
extended with these phrasings, and the pooled scores leave out the two trials of the repeated models
that ended this way.
\\
\textbf{2. Enforce restrictions outside the agent's privilege domain.} Studio's guardrail is the
user's approval of shell commands and of writes outside the workspace. With it on, the nine
permission restores in the case study would have stopped at an approval prompt, but reads are not
gated, and the other four leaks were reads. An unattended evaluation must turn approval off, so it needs a boundary of its
own. Without one, an agent's process also emptied the home directory during a sweep, deleting the
harness and 13 hours of results, and an agent's test script made a fixture unreadable and failed the
next 30 trials.
\\
\textbf{3. Measure every skill change on more than one model.} The same change helped one model
substantially, helped a second partly, and left a third unchanged.

\section{Related Work}
Kapoor et al.~\cite{kapoor2024agents} and Zhu et al.~\cite{zhu2025abc} describe
practices for rigorous agent evaluation. These approaches isolate the agent from the scorer.
In contrast, we evaluate agents within the operational gateway to capture their interactions with the deployed system, which introduces additional challenges for experimental isolation.
Frontier models have been observed modifying tests to
raise their scores~\cite{metr2025rewardhacking}. In our study, agents circumvented an experimental restriction while pursuing their assigned tasks, rather than explicitly attempting to improve their evaluation scores.

\section{Limitations and Conclusion}

Eighteen of the 21 open-weight models ran each task once, so we report a top group and do not rank
within it. The repeated trials cover only the build tasks of four models, and they used the extended
question detector, which the baseline did not. The case study covers four models and four tasks and is scored on
L0 only. The skill changes were measured only on individual models during screening. During the trials, the container build
failed in all 33 baseline trials that reached it, because the evaluation host lacked the user namespace
that Apptainer needs, and the harness does not request a plan after a failed build. We later replayed
these stages on the stored projects with user namespaces enabled, without calling a
model. In the replay, 27 of the 33 builds succeeded, and the planning counts in RQ1 come from it. The L0 and L1 scores predate the replay, so L0 understates
every model on the seven tasks graded through build or plan.

We showed how skills written with a frontier model can be adapted for open-weight models and evaluated within the science gateway in which they are deployed.
In repeated trials, the best of the three open-weight models we repeated
trailed Claude by 0.06 on workflow building. After one trial per task, the leading models differed far more in speed than in
quality, and they failed where a skill instruction allowed more than one reading. 
The evaluation also revealed that failures in the harness can be mistaken for model failures, and that agents can alter experimental conditions when those conditions are within their control. 
In a controlled ablation, an initially significant effect disappeared when the experimental restriction was enforced outside the agents' privilege domain. These findings highlight the need to audit evaluation infrastructure and enforce experimental controls independently of the agents being evaluated. 
Future work includes
repeated trials for every model and a comparison of the specialist agents with the default agent.

\textbf{Artifact availability.} The harness, the tasks, the per-trial records, and the analysis scripts are publicly available~\cite{promptlab_artifact}.

\section*{Acknowledgments}
This work is supported by Pegasus AI NSF grant \#2513101 and the FABRIC Testbed (NSF \#2330891).

\bibliographystyle{IEEEtran}
\bibliography{refs}

\end{document}